\documentclass[letterpaper,10pt,journal,twoside]{IEEEtran}

\usepackage{amsmath,amssymb,amsfonts}
\usepackage{graphicx}
\usepackage{booktabs}
\usepackage{cite}
\usepackage{url}
\usepackage[hidelinks]{hyperref}
\usepackage{microtype}
\hypersetup{colorlinks=true, linkcolor=blue, urlcolor=blue}

\title{TacSE3: Equivariant SE(3) Motion Estimation from Low-Texture Visuotactile Images for In-Gripper Tracking and Compensation}

\author{
Zhongyuan Liao,
Junzhe Wang,
Qingyang Liu,
Zhenmin Huang,
Jun Ma,~\IEEEmembership{Senior Member,~IEEE},
Yi Cai,
Fei Meng*,
Haobo Liang*, and
Michael Yu Wang*,~\IEEEmembership{Fellow,~IEEE}
\thanks{
Zhongyuan Liao, Zhenmin Huang, Fei Meng, and Haobo Liang are with the Hong Kong Center for Construction Robotics, 
The Hong Kong University of Science and Technology, Hong Kong, China }

\thanks{
Junzhe Wang, Qingyang Liu, Jun Ma, and Yi Cai are with The Hong Kong University of Science and Technology (Guangzhou), China.
}

\thanks{
Michael Yu Wang is with the School of Engineering, 
Great Bay University, Dongguan, China.
}

\thanks{*Corresponding authors: Fei Meng (feimeng@ust.hk), Haobo Liang (hbliang@ust.hk), and Michael Yu Wang (mywang@gbu.edu.cn).}

}

\date{\today}

\begin{document}
\maketitle

\begin{abstract}

Robotic in-hand manipulation requires reliable object-motion tracking under frequent visual occlusion, yet low-texture visuotactile images provide few stable correspondences for conventional image- or geometry-matching methods. This paper presents TacSE3, a tactile motion-estimation pipeline that converts low-texture visuotactile observations into a decoupled three-dimensional force field and estimates incremental rigid-body motion on $SE(3)$. The method derives planar translation from contact-centroid motion and estimates rotation primarily from shear-related tactile responses, yielding a physically interpretable signal for in-gripper tracking and compensation. Experiments with paired DM-Tac fingertip sensors show that dual-sensor sensing reduces translation--rotation ambiguity, supports rotation tracking across axes and object geometries, and provides a lightweight compensation signal that improves disturbance tolerance in downstream manipulation tasks without retraining the base policy.

\textbf{Note to Practitioners}---This paper targets in-gripper manipulation settings where vision is unreliable because of occlusion. For textureless or symmetric objects, Vision-based tactile sensors (VBTS) based tracking is often unstable because salient visual features are scarce and rotational motion often produces only subtle appearance changes. Our method avoids relying on texture matching by converting low-texture visuotactile images into a decoupled three-dimensional force field and estimating local motion on $SE(3)$ from that field. In practice, it is best used as a lightweight compensation module that can be attached to an existing manipulation policy when vision becomes unreliable, especially for VBTS sensors with weak texture response and for smooth or symmetric textureless objects. Project website: \url{https://yuanzero.github.io/tacSE3/}.
\end{abstract}


\section{Introduction}
\label{sec:introduction}

Touch plays a fundamental role in human perception and manipulation~\cite{lederman87}. Through touch, humans can infer local geometry, pose variation, slip, and object properties even when vision is unavailable or unreliable. Endowing robots with comparable contact-based spatial awareness remains challenging, yet it is increasingly important for contact-rich manipulation, where sensing at the contact interface often determines whether a task succeeds or fails.

Vision-based tactile sensors (VBTS), such as GelSight~\cite{yuan2017gelsight}, Deltact~\cite{zhang_deltact_2022}, and DIGIT~\cite{lambeta2020digit} have enabled high-resolution observation of contact, allowing robotic systems to sense local geometry, deformation, and slip directly at the contact interface. Relative to external vision, VBTS sensing remains informative under occlusion and is naturally aligned with contact-rich interaction. These advantages have motivated substantial interest in tactile perception for object pose estimation, in-hand manipulation, and reactive control~\cite{bimbo16,huang24,suresh24}. Historically, many successful VBTS pipelines have benefited from sensors whose raw tactile images contain rich texture, embedded markers, or photometrically distinctive appearance changes, making raw-image registration and tactile reconstruction more tractable.

Despite this progress, tactile perception has not yet achieved the level of generality and robustness that vision and LiDAR provide for spatial perception. In practice, tactile sensing often remains a supporting modality that refines visual estimates or compensates for occlusion, rather than serving as a primary source of geometric state estimation~\cite{suresh24,li25}. Standalone tactile systems for tracking or reconstruction typically remain less robust and less scalable than their visual counterparts~\cite{sodhi22,wang18,newcombe11,li23}. A central reason is that tactile sensing is inherently local: each contact frame is information-rich but spatially limited, and establishing reliable spatial relations across frames is difficult without strong geometric cues or external context.

This limitation is particularly severe for low-texture VBTSs. Many existing motion-estimation pipelines rely, either explicitly or implicitly, on salient texture, marker motion, dense optical flow, or stable image correspondences. Under weak-texture sensing conditions, however, these cues become fragile. For smooth and texture-poor objects, such as a polished sphere, the tactile image provides few repeatable appearance cues, so even substantial object rotation may induce only subtle image variation. Small appearance changes therefore need not correspond to physically meaningful motion, and local image measurements may be dominated by illumination artifacts, elastic deformation, or contact-area variation. As a result, methods that reconstruct state directly from raw tactile images become least reliable precisely in the regime where robust contact-centric estimation is most needed. TacSE3 is designed for that setting: instead of relying on texture, it reconstructs $SE(3)$ motion from a decoupled 3D force field.

In-gripper manipulation particularly highlights this challenge. Securely grasped objects frequently undergo subtle translations or rolling motions during execution. While often imperceptible to global vision, these micro-motions accumulate, leading to tracking drift and degraded grasp stability. What is needed in such settings is not merely a contact detector or task-specific latent feature, but a tactile output that is geometrically meaningful and directly usable by downstream control.

This paper addresses the following question: how can one construct a reliable $SE(3)$ motion estimate directly from tactile images when explicit texture tracking is unreliable? Our answer, embodied in TacSE3, is to treat the tactile image primarily as a geometric measurement of contact rather than as a conventional textured image. Instead of reconstructing motion from raw appearance, we first decouple the tactile response into a 3D force field with tangential and normal components, and then recover rigid-body motion from that field through an $SE(3)$ geometric model designed to keep the estimated motion consistent in the end-effector frame across changing arm poses and contact locations. The resulting estimate is physically interpretable, temporally consistent, and readily integrated into downstream robotic systems.

The scope of the contribution is deliberately focused. The goal is not to develop a fully equivariant visuotactile architecture, nor to impose geometric equivariance on the visual branch. The notion of equivariance considered here is specific to the tactile-to-$SE(3)$ mapping: the estimated motion should remain consistent when expressed in the end-effector frame even if the robot pose or local contact location changes. The main contribution therefore lies on the tactile side: a low-texture tactile image stream is converted into a decoupled 3D force field, and that field is then mapped to an $SE(3)$ motion estimate that captures local rigid-body change at the contact interface. This output can be used either as an online tracking signal for compensation or as a lightweight post-processing signal that improves the adaptability of an existing control pipeline. In this sense, the paper centers on force-field-to-$SE(3)$ reconstruction as a geometric interface between VBTS sensing and control.

The main contributions are as follows:
\begin{itemize}
    \item We formulate tactile-image-to-$SE(3)$ estimation for low-texture visuotactile sensing as a geometric state estimation problem centered on local rigid-body motion at the contact interface.
    \item We propose a tactile motion estimation pipeline that decouples raw tactile observations into a 3D force field and reconstructs rigid-body twist through a physically grounded $SE(3)$ model.
    \item We evaluate the estimated rotational signal on a grasped ball under controlled rotations about $r_x$, $r_y$, and $r_z$.
    \item We show that the estimated rotation can be attached as a post-processing signal to improve robot adaptability when a grasped object rotates inside the gripper.
\end{itemize}

The rest of the paper is organized as follows. Section~\ref{sec:related} reviews related work in visuotactile sensing, tactile motion estimation, and geometric rigid-body modeling. Section~\ref{sec:problem} formulates the problem. Section~\ref{sec:method} presents the proposed force-field-to-$SE(3)$ estimation method. Section~\ref{sec:experiments_discussion} describes the experimental setup, reports the quantitative and policy-level results, and discusses the main findings. Section~\ref{sec:conclusion} concludes the paper.

\section{Related Work}
\label{sec:related}

\subsection{VBTS-Based Perception for Manipulation}
VBTSs have become an important sensing modality for contact-rich manipulation because they provide dense local measurements of contact geometry and deformation~\cite{li2025classification,he2025survey}. Prior work has used VBTS observations for surface-normal estimation, object pose estimation, and in-hand manipulation~\cite{li2018end,bauza23,huang24}. Many such systems operate directly on raw tactile images, normal maps, marker motion, or image correspondences, which is effective when the tactile image contains sufficiently rich texture or otherwise distinctive visual structure. More recent visuotactile systems combine tactile sensing with vision to improve in-hand perception under occlusion~\cite{dikhale2022visuotactile,suresh24,li25}. These studies have established the practical value of tactile sensing, but many methods still rely on strong image cues, object-specific priors, or global context from vision. Prior studies have begun to explore decomposed 3D force representations derived from VBTS observations for downstream perception and manipulation~\cite{du2024hanging,liao2025quantitative}, but such force-structured representations remain less common in tactile motion estimation.

\subsection{Tactile Object Pose Estimation and Tracking}
Object pose estimation from touch has been studied in both model-based and model-free settings. When object geometry is known, tactile observations can be matched against local object geometry using covariance statistics, particle filters, or learned object-specific embeddings~\cite{bimbo16,suresh22midas,bauza23}. For previously unseen objects, prior work has commonly relied on tactile-derived point clouds, normal maps, or local surface descriptors. For example, ICP-style registration has been used to estimate frame-to-frame object motion from tactile point clouds~\cite{wang21_1,chen91}, while PatchGraph combines local tactile registration with factor-graph reasoning for in-hand tracking~\cite{sodhi22}. NormalFlow shows that normal-map representations can provide highly accurate 6-DoF tracking~\cite{huang24}, but the authors also report difficulties on highly symmetric objects, such as balls and cylinders, because matching normal maps or point clouds becomes ambiguous. More broadly, these methods reconstruct motion largely from raw geometric or image-level observations. Optical-flow, marker-tracking, ICP, Procrustes, and filtering-based baselines are therefore highly relevant when repeatable visual features, dense point correspondences, or reliable geometry are available. The present work focuses on the complementary low-texture in-gripper regime, where such correspondences are weak or ambiguous, and introduces an explicit intermediate representation, namely a decoupled 3D force field, from which the $SE(3)$ motion is subsequently reconstructed. This design is intended to reduce the dependence on correspondence quality in texture-poor and symmetry-prone settings rather than to replace full object-level tactile tracking pipelines in all regimes.

However, long-horizon tactile tracking on novel objects remains difficult, as individual touches yield strictly local geometric constraints. Many systems address this limitation by incorporating vision or other global cues~\cite{dikhale2022visuotactile,suresh24,li25}. Our setting is different. We do not seek full object-level state estimation through visuotactile fusion; instead, we focus on recovering local rigid-body motion directly from tactile images and using that signal as a control-facing estimate of relative in-gripper motion.


\subsection{Rigid-Body Estimation and Tactile Control}
Rigid-body motion estimation in $SE(3)$ is a standard formulation in robotics and geometric vision because it enforces physically meaningful constraints on observed motion. In parallel, geometric priors and equivariant representations have become increasingly important in robot learning and manipulation~\cite{thomas2018tensor,ryu2023diffusionedfs,raven2024}.

Recent work has further explored $SE(3)$-aware policy design and equivariant manipulation models~\cite{seo2025equicontact,zhu2025equact,zhu2025residual,gao2025riemann}. The present work is narrower in scope but aligned in spirit: rather than seeking full multimodal equivariance, we impose a rigid-body geometric prior specifically on the tactile branch so that the output remains physically interpretable and stable under local pose variation.

Tactile feedback has recently become central to a broad range of manipulation and control problems, including tactile servoing, visuotactile policy learning, in-hand rotation, and non-prehensile control under occlusion~\cite{simshear25,3dvitac25,mimictouch25,text2touch25,anyrotate25,visuotactilecontrol25}. Recent systems also highlight the importance of tactile sensing for data collection, such as universal manipulation interface based on the VBTS ~\cite{cheng2026tacumi,exumi25}, representation learning, and large-area contact awareness in dexterous manipulation~\cite{kinedex25,tacx25,dexskin25}.

Our emphasis differs from most prior work. Rather than learning an invariant latent representation or designing a task-specific tactile controller, we first reconstruct a decoupled 3D force field from tactile observations and then estimate a continuous $SE(3)$ motion signal from that field. This choice is motivated by low-texture VBTS settings, in which force-structured geometric inference can be more informative than a correspondence-heavy or purely raw-image-driven representation.

\section{Problem Formulation}
\label{sec:problem}

We consider a robotic gripper equipped with a low-texture VBTS mounted at or near the fingertip. At each time step $t$, the sensor produces a tactile observation
\begin{equation}
\mathcal{O}_t = \{ \mathcal{I}_t, \mathcal{D}_t, \mathcal{C}_t \},
\end{equation}
where $\mathcal{I}_t$ denotes the tactile image, $\mathcal{D}_t$ denotes a depth-like or reconstructed surface signal when available, and $\mathcal{C}_t$ denotes any additional contact cue derived from the sensor, such as deformation magnitude or confidence. Depending on the hardware, some entries may be measured directly, whereas others may be obtained through preprocessing.

In our implementation, the tactile sensor is the DM-Tac VBTS from Daimon Robotics, mounted on the robotic gripper. The sensor captures tactile signals at 120 Hz with an image resolution of $320 \times 240$. During grasping, the sensor produces deformation measurements that can be decomposed into normal and shear force distributions. These force-related quantities form the basis of the intermediate representation used in this work.

Let an object remain in contact with the sensor surface over a short temporal window and undergo a small rigid-body motion relative to the gripper. The objective is to estimate this relative motion as an element of $SE(3)$:
\begin{equation}
T_t =
\begin{bmatrix}
R_t & \mathbf{p}_t \\
\mathbf{0} & 1
\end{bmatrix}
\in SE(3),
\end{equation}
where $R_t \in SO(3)$ denotes rotation and $\mathbf{p}_t \in \mathbb{R}^3$ denotes translation. In the gripper setting considered here, the dominant translational component is the in-plane motion of the contact patch, while the normal component is inferred from sensor deformation and treated as negligible for object-level compensation because the gripper mechanically constrains motion along the sensor normal.

The central challenge is that $\mathcal{I}_t$ does not contain sufficiently strong texture to support reliable direct correspondence matching. We therefore seek a model that infers rigid-body motion from contact geometry and temporal tactile variation rather than from image texture alone. More concretely, given a short sequence of tactile observations $\mathcal{O}_{t-k:t}$, we seek an estimator
\begin{equation}
\Phi: \mathcal{O}_{t-k:t} \mapsto \hat{T}_t,
\end{equation}
that is accurate, temporally stable, and physically interpretable.

The estimated motion must satisfy two practical requirements: it should be accurate and stable enough for online in-gripper tracking and compensation, and its rotation output should remain useful even when detached from the full tracking pipeline and used as a lightweight post-processing signal.

An additional practical difficulty arises when the estimate is derived from only a single VBTS. In that case, translation and rotation can induce highly similar local deformations on the sensor surface, which makes the inferred motion ambiguous and degrades estimation accuracy. This issue is especially evident for compact contact patches and texture-poor objects. Using two VBTSs mounted on opposing gripper fingers provides complementary observations of the same object motion. The two contact streams can therefore cross-validate one another and reduce the ambiguity between translation-induced and rotation-induced deformation.

\begin{figure*}[t]
    \centering
    \includegraphics[width=1\linewidth]{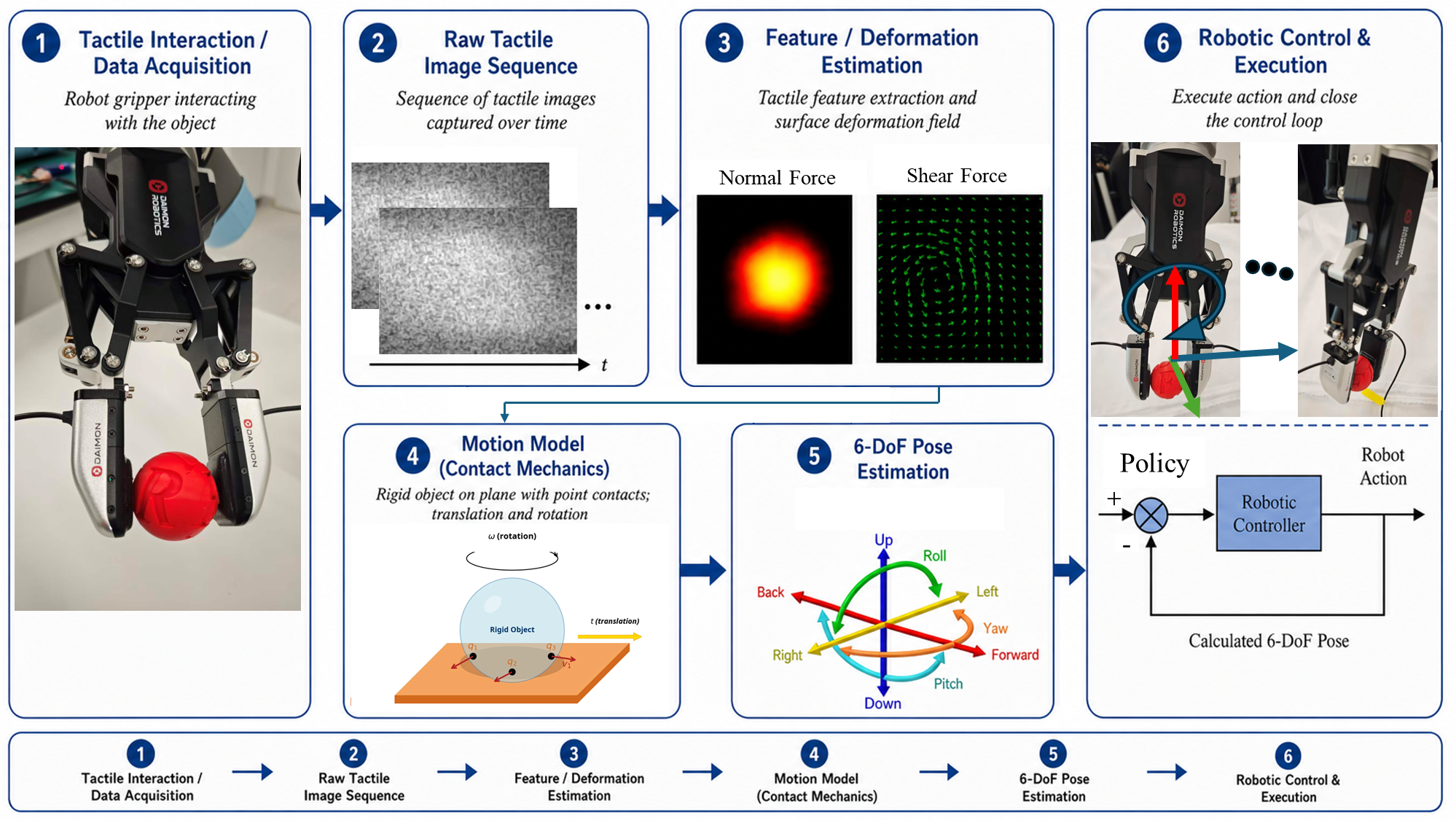}
    \caption{Overview of the proposed tactile-image-to-$SE(3)$ estimation pipeline.}
    \label{fig:pipeline}
\end{figure*}

Fig.~\ref{fig:pipeline} summarizes the proposed pipeline. Given a sequence of tactile observations, we first identify the active contact region and reconstruct a local 3D representation of the contact surface. We then decouple the tactile response into a 3D force field with tangential and normal components and use this field to construct local motion constraints. These constraints are fit with a rigid-body model in $SE(3)$. Finally, the estimated twist is temporally filtered and integrated to produce a stable motion output that can be used either within an online compensation loop or as a post-processing signal for adaptive manipulation.

The method is designed for low-texture sensing conditions. Rather than depending on explicit tactile texture or marker tracks, it exploits geometric structure that remains available even when the image appearance is only weakly informative. The design follows three principles: contact-first sensing, force-field decoupling, and rigid-body consistency. The tactile stream is first converted into contact geometry, then into a decoupled 3D force field, and finally constrained by rigid-body kinematics to produce an interpretable $SE(3)$ estimate.

\section{SE(3) Construction from Tactile Images}
\label{sec:method}

\subsection{Low-Texture Tactile Image Representation}
\label{subsec:low_texture}

The first design principle of the method is to avoid explicit dependence on salient texture. In low-texture visuotactile sensing, image patches often lack repeatable keypoints, and local intensity changes may not correspond reliably to physical motion. Marker-free regions may exhibit smooth intensity fields, while local optical changes may be dominated by illumination variation or elastic deformation rather than by object motion. Many earlier VBTS pipelines operate directly on raw tactile images or raw-image reconstructions because their sensing conditions provide sufficiently rich texture to support such processing. In the present setting, that assumption is weak. Nevertheless, the tactile image still contains useful information in the form of contact shape, deformation distribution, boundary evolution, and depth variation over time.

We therefore treat the tactile image as a geometric measurement of contact rather than as a conventional textured image. This shift in viewpoint is central to the paper. The objective is not to track image features directly, but to infer how the contact geometry evolves over time, decouple the tactile response into physically meaningful tangential and normal components, and explain that evolution through a rigid-body motion model. This perspective motivates the construction of 3D contact points and the subsequent decoupled 3D force field.

\subsection{Decoupled 3D Force Field and Contact Point Construction}
\label{subsec:force_contact}

A representative dense color-pattern VBTS pipeline~\cite{zhang_deltact_2022} turns raw tactile images into a dense 2D displacement field by tracking the printed pattern with optical flow, using an adaptive reference frame when deformation is large.
Let $\mathbf{m}$ denote a point in image coordinates. At each $\mathbf{m}$, a displacement $\mathbf{u}(\mathbf{m})$ is obtained by minimizing the squared photometric error between a reference image $I_{\mathrm{ref}}$ and the current image $I_t$ over a window $\mathcal{W}(\mathbf{m})$,
\begin{equation}
\mathbf{u}^\star(\mathbf{m}) =
\arg\min_{\mathbf{u}'}
\sum_{\mathbf{s}\in\mathcal{W}(\mathbf{m})}
\left[
I_t\!\left(\mathbf{s}+\mathbf{u}'\right) - I_{\mathrm{ref}}(\mathbf{s})
\right]^2 .
\label{eq:dense_flow}
\end{equation}
The resulting dense planar flow $\mathbf{w}(\mathbf{m})=\mathbf{u}^\star(\mathbf{m})$ is smoothed with an isotropic Gaussian kernel of covariance $\boldsymbol{\Sigma}_{\sigma}=\sigma^2\mathbf{I}_2$; the accumulated \emph{Gaussian density} map $\mathcal{G}(\mathbf{m})$ captures local expansion of $\mathbf{w}$ and yields a shift-invariant proxy for out-of-plane deformation, with relative height taken proportional to $-\mathcal{G}(\mathbf{m})$.
To separate in-plane shear from the component associated with normal contact, $\mathbf{w}$ is decomposed by natural Helmholtz--Hodge decomposition (NHHD) as
\begin{equation}
\mathbf{w}(\mathbf{m}) =
\mathbf{d}(\mathbf{m}) + \mathbf{r}(\mathbf{m}) + \mathbf{k}(\mathbf{m}),
\label{eq:nhhd}
\end{equation}
where $\mathbf{d}$ is curl-free ($\nabla\times\mathbf{d}=\mathbf{0}$), $\mathbf{r}$ is divergence-free ($\nabla\cdot\mathbf{r}=0$), and $\mathbf{k}$ is harmonic ($\nabla\times\mathbf{k}=\mathbf{0}$, $\nabla\cdot\mathbf{k}=0$); calibrated linear maps from $\mathcal{G}$, $\mathbf{d}$, $\mathbf{r}$, and $\mathbf{k}$ then yield spatial distributions of normal and shear force~\cite{zhang_deltact_2022}.
The per-pixel stack below abstracts this image-to-force structure for the estimator used in this work.

Given two consecutive tactile observations, a decoupled 3D force field is first constructed to capture the local contact response. For each pixel, the tactile response is decomposed into tangential and normal components,

\begin{equation}
\mathbf{f}(x,y) =
\begin{bmatrix}
f^x(x,y) \\
f^y(x,y) \\
f^z(x,y)
\end{bmatrix},
\end{equation}

where $f^x$ and $f^y$ denote tangential responses, and $f^z$ represents the normal response inferred from temporal variation in the reconstructed contact geometry.

\begin{figure}
    \centering
    \includegraphics[width=1\linewidth]{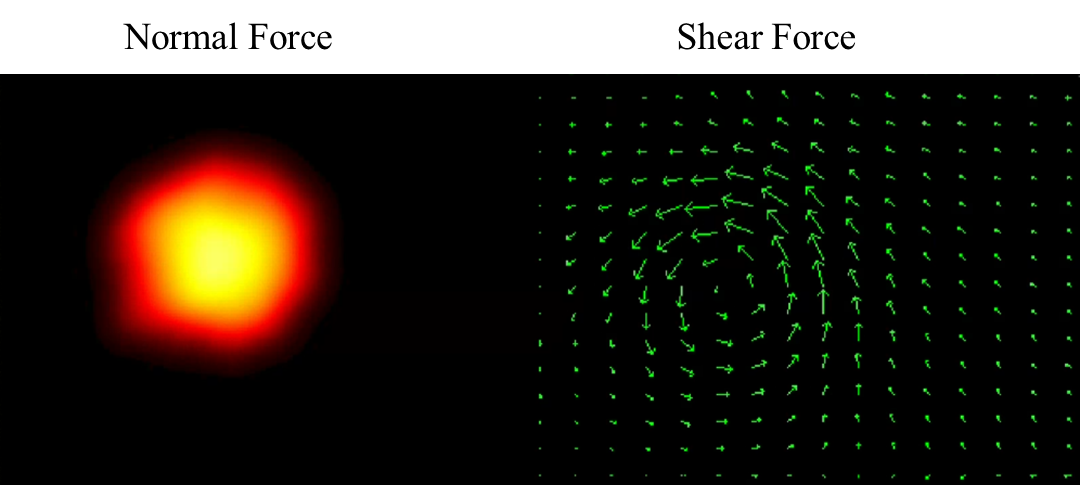}
    \caption{Decoupled tangential and normal responses derived from tactile deformation.}
    \label{fig:force}
\end{figure}

Fig.~\ref{fig:force} illustrates the deformation of the tactile surface under contact. The displacement between the undeformed and deformed states produces a dense response field, which can be decomposed into tangential and normal components following prior work on dense-pattern VBTS sensing and 3D reconstruction~\cite{zhang_deltact_2022,du_3d_2022}.

To localize the active contact region, we extract pixels with significant normal response. Specifically, a filter function is applied to the normal field,

\begin{equation}
\Omega_c = \{ (x,y) \mid f_{\text{filter}}(f^z(x,y)) > \tau \},
\end{equation}

where $\tau$ denotes the threshold for valid contact detection. This step identifies the reliable contact patch while suppressing noise and marginal deformation.

For each valid pixel $(i,j) \in \Omega_c$, a 3D point is constructed in the local tactile coordinate frame. Let $\kappa_x$ and $\kappa_y$ denote the spatial scale factors of the sensor in the tangential directions. The coordinates are defined as

\begin{equation}
x_{ij} = \frac{i-i_0}{\kappa_x}, \qquad
y_{ij} = \frac{j-j_0}{\kappa_y},
\end{equation}

The normal coordinate is estimated as a monotonic function of the normal response, using the undeformed tactile sensor plane as a reference:

\begin{equation}
z_{ij} = g(f^z_{ij}),
\end{equation}
where $g(\cdot)$ is a monotonically increasing function mapping the normal response $f^z_{ij}$ to an estimate of local deformation along the sensor normal. In this formulation, the undeformed sensor plane corresponds to $z_{ij}=0$, and larger $f^z_{ij}$ values indicate greater out-of-plane displacement. This term provides the local $z$-direction displacement induced by normal deformation of the tactile surface. However, in our in-gripper setting, this normal displacement is much smaller than the tangential $x$--$y$ motion and is further constrained by the opposing-finger gripper structure. We therefore use $z_{ij}$ to construct the local 3D contact geometry, but treat the net object-level translation along the sensor normal as negligible in the subsequent compensation analysis.

Each valid pixel is therefore mapped to a 3D contact point

\begin{equation}
\mathbf{q}_{ij} =
\begin{bmatrix}
x_{ij} \\
y_{ij} \\
z_{ij}
\end{bmatrix}.
\end{equation}

The resulting set of points provides a dense and reliable 3D representation of the contact surface. This formulation converts raw tactile observations into a physically structured representation suitable for subsequent rigid-body motion estimation.

\subsection{Rigid-Body Motion Model in \texorpdfstring{$SE(3)$}{SE(3)}}
\label{subsec:rigid_model}

\begin{figure}
    \centering
    \includegraphics[width=0.75\linewidth]{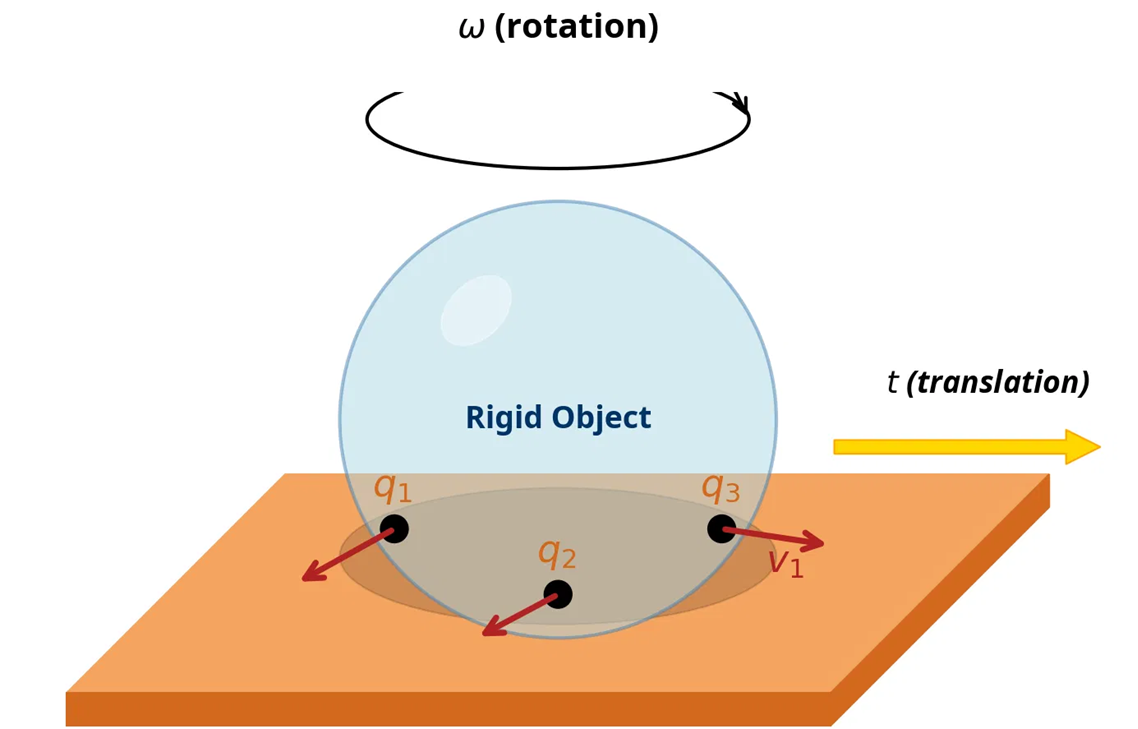}
    \caption{Rigid-Body Motion Model.}
    \label{fig:rigidmodel}
\end{figure}

The decoupled 3D field introduced above is not interpreted here as a direct measurement of accumulated pose. Instead, it is treated as a local motion-related response defined over the contact patch. This choice is justified by the short-horizon contact model adopted in this paper: over sufficiently small sampling intervals, and in the absence of severe slip or gross non-rigid deformation, the deformation of the compliant sensing surface is locally coupled to the relative velocity of the object at the contact interface, shown in Fig.~\ref{fig:rigidmodel}. 

Consequently, after calibration and normalization, the three response components can be used to construct a local velocity proxy
\begin{equation}
\tilde{\mathbf{v}}_i =
\begin{bmatrix}
\tilde{v}^x_i \\
\tilde{v}^y_i \\
\tilde{v}^z_i
\end{bmatrix},
\end{equation}
which approximates the instantaneous motion of the contact point $\mathbf{q}_i$ during the sampling interval. The key point is that the estimator does not require an exact force-to-velocity equality; it requires only that the calibrated responses preserve the local direction and relative magnitude of contact motion sufficiently well to define kinematic constraints. In this sense, the tactile field serves as an intermediate representation that links the measured contact response to a rigid-body kinematic model.

We then model this local velocity proxy with a rigid-body twist:
\begin{equation}
\mathbf{v}_i = \boldsymbol{\omega} \times \mathbf{q}_i + \mathbf{t},
\end{equation}
where $\boldsymbol{\omega} \in \mathbb{R}^3$ is the angular velocity and $\mathbf{t} \in \mathbb{R}^3$ is the translational velocity. Importantly, $\boldsymbol{\omega}$ is an instantaneous rotational rate rather than an accumulated rotation angle. Defining
\begin{equation}
\boldsymbol{\xi} =
\begin{bmatrix}
\boldsymbol{\omega} \\
\mathbf{t}
\end{bmatrix}
\in \mathbb{R}^6,
\end{equation}
we obtain a standard linear relationship between the reconstructed local motion proxy and the rigid-body twist. The corresponding rotation increment over one sampling interval is therefore determined by $\boldsymbol{\omega}\Delta t$, whereas the cumulative orientation is obtained only after temporal integration, as described in the next subsection.

For each contact point $\mathbf{q}_i = (x_i, y_i, z_i)^T$, the expanded equations are
\begin{equation}
\begin{aligned}
\tilde{v}^x_i &= \omega_y z_i - \omega_z y_i + t_x, \\
\tilde{v}^y_i &= \omega_z x_i - \omega_x z_i + t_y, \\
\tilde{v}^z_i &= \omega_x y_i - \omega_y x_i + t_z.
\end{aligned}
\end{equation}
Due to the constrained planar contact, motion along the normal direction is typically small and can be neglected. Therefore, the third equation 
$\tilde{v}^z_i = \omega_x y_i - \omega_y x_i + t_z$
can be omitted, and both $\tilde{v}^z_i$ and $t_z$ are ignored. Only the in-plane motion constraints are retained for estimation.
Each point contributes a linear constraint of the form
\begin{equation}
\mathbf{A}_i \boldsymbol{\xi} = \mathbf{b}_i,
\end{equation}
where
\begin{equation}
\mathbf{A}_i =
\begin{bmatrix}
0 & z_i & -y_i & 1 & 0 & 0 \\
-z_i & 0 & x_i & 0 & 1 & 0 \\
y_i & -x_i & 0 & 0 & 0 & 1
\end{bmatrix},
\qquad
\mathbf{b}_i =
\begin{bmatrix}
\tilde{v}^x_i \\
\tilde{v}^y_i \\
\tilde{v}^z_i
\end{bmatrix}.
\end{equation}
Stacking all valid points yields an overdetermined linear system of the form
\begin{equation}
\mathbf{A}\boldsymbol{\xi} = \mathbf{b}.
\end{equation}

This formulation is particularly attractive for low-texture sensing because it ties the estimate to a rigid-body prior rather than to a texture-rich image-formation model. The decoupled 3D force field absorbs much of the ambiguity in the raw tactile image, while the subsequent $SE(3)$ fit constrains the final estimate through rigid-body geometry. Even when the reconstructed motion proxy is noisy, the global twist remains constrained by the structure of rigid motion. Although the present implementation is developed for VBTS sensing, the rigid-body modeling itself is not specific to a particular tactile hardware platform. More generally, the same $SE(3)$ motion model can be applied to any tactile sensor that provides a calibrated three-dimensional force field or an equivalent local contact-motion proxy over the contact patch. In that sense, VBTS is one practical realization of the sensing front end, whereas the rigid-body reconstruction step is broadly compatible with other tactile sensing modalities that can produce comparable 3D contact-field information.

\subsection{Least-Squares \texorpdfstring{$SE(3)$}{SE(3)} Estimation and Temporal Smoothing}
\label{subsec:estimate}

The rigid-body twist is estimated at each time step by minimizing the residual error over all contact points:
\begin{equation}
\boldsymbol{\xi}^{*} = \arg\min_{\boldsymbol{\xi}} \| \mathbf{A}\boldsymbol{\xi} - \mathbf{b} \|_2^2.
\label{eq:twist_ls}
\end{equation}
This least-squares formulation is computationally efficient for the typical number of valid contact points and provides a direct estimate of the instantaneous rigid-body twist at each time step.

\textbf{Contact-centroid planar translation.}
In our implementation, planar translation is \emph{not} taken as the constant $(t_x,t_y)$ obtained by jointly least-squares fitting the shear-induced velocity field together with rotation in~Eq.\eqref{eq:twist_ls} (nor, when a depth twist is used, as the corresponding normal component $t_z$). Instead, we compute the centroid of the active \emph{contact mask} on the sensor plane and define $(\Delta x,\Delta y)$ from the planar displacement of that centroid over time.
Rotation is estimated separately from shear-based motion constraints. By fitting only the rotational part of the rigid-body model, we prevent translation from being co-estimated with $\boldsymbol{\omega}$ in the same least-squares problem, thereby avoiding competition with the centroid-based planar motion.
This choice has two advantages: the planar displacement has a clear physical interpretation as motion of the contact patch center on the sensor, rather than an abstract uniform translation term in the velocity-field model; and it decouples translation from rotation, avoiding the coupling and ill-conditioning that arise when $\boldsymbol{\omega}$ and $\mathbf{t}$ are co-estimated in the same overdetermined system, which yields cleaner rotation estimates.

When this decoupling is used, the translational components in the integrated pose are the centroid-based $(\Delta x,\Delta y)$ (with $t_z=0$ under the planar-contact approximation), while the rotational part of $\boldsymbol{\xi}$ comes from the shear-only least-squares solution.

To reduce frame-to-frame jitter caused by sensing noise, we apply temporal averaging to the estimated twist sequence. Let $\boldsymbol{\xi}^{*}_t$ denote the estimate at time step $t$. The smoothed estimate is computed as
\begin{equation}
\bar{\boldsymbol{\xi}}_t = \frac{1}{L}\sum_{j=t-L+1}^{t}\boldsymbol{\xi}^{*}_j,
\end{equation}
where $L$ is the averaging window length. In practice, this smoothing step suppresses high-frequency fluctuations while preserving the dominant motion trend required for tracking and compensation.

The estimated twist is first embedded into the Lie algebra of $SE(3)$ as

\begin{equation}
\hat{\bar{\boldsymbol{\xi}}} =
\begin{bmatrix}
[\boldsymbol{\omega}]_\times & \mathbf{t} \\
\mathbf{0} & 0
\end{bmatrix}
\in \mathfrak{se}(3),
\end{equation}

where $\boldsymbol{\omega} \in \mathbb{R}^3$ and $\mathbf{t} \in \mathbb{R}^3$ denote the angular and translational components of the rigid-body motion, respectively, and $[\boldsymbol{\omega}]_\times$ is the skew-symmetric matrix representation of angular velocity.

The pose is then updated on the $SE(3)$ manifold using the exponential map:

\begin{equation}
T_{t+1} = T_t \exp\left(\hat{\bar{\boldsymbol{\xi}}} \Delta t\right),
\end{equation}
where $\exp(\cdot)$ maps the twist from the tangent space $\mathfrak{se}(3)$ to the Lie group $SE(3)$. 
This formulation enables incremental pose updates through a sequence of small rigid-body motions. When quantitative evaluation against real $SE(3)$ pose data is required, the integrated tactile pose is further aligned to the ground-truth pose frame through a calibration mapping on $SE(3)$, so that the estimated local contact motion can be consistently compared with the real object pose despite the nonlinear relationship between tactile measurements and global pose.

\begin{figure}
    \centering
    \includegraphics[width=1\linewidth]{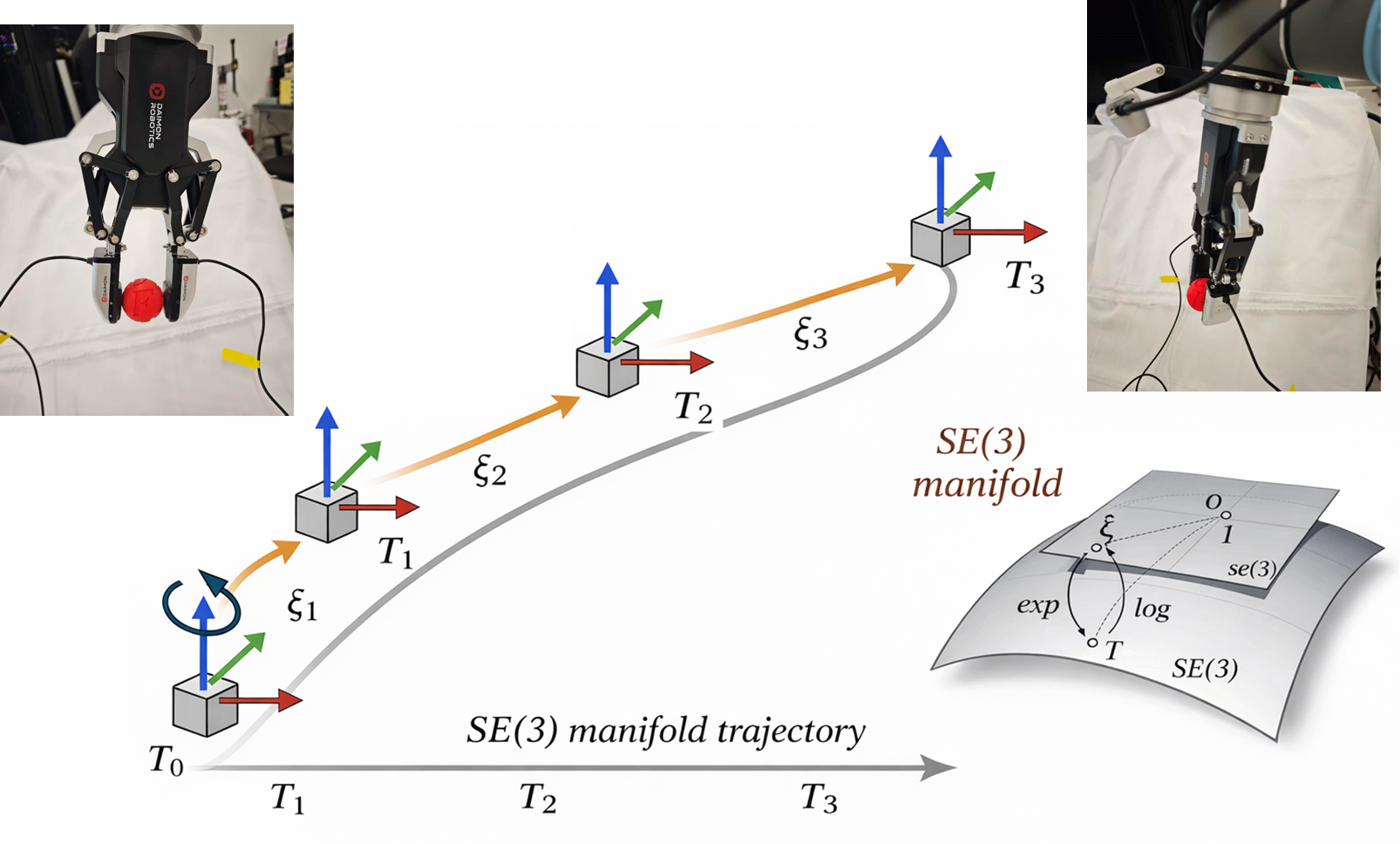}
    \caption{
    Pose tracking on the $SE(3)$ manifold. 
    Small twists $\boldsymbol{\xi}_t \in \mathfrak{se}(3)$ are mapped to $SE(3)$ via the exponential map and sequentially integrated to form a continuous pose trajectory 
    $T_0 \rightarrow T_1 \rightarrow T_2 \rightarrow T_3$.
    }
    \label{fig:SE3motion}
\end{figure}

As illustrated in Fig.~\ref{fig:SE3motion}, each estimated twist corresponds to a local motion on the tangent space, which is then projected onto the $SE(3)$ manifold via the exponential map. The accumulated poses therefore form a continuous trajectory 
\(
T_0 \rightarrow T_1 \rightarrow T_2 \rightarrow T_3
\)
on the $SE(3)$ manifold, where each step represents a small rigid-body transformation.

Rather than directly predicting absolute orientation from frame pairs, we obtain the rotation component of $T_t$ by integrating angular velocity over time. This approach ensures that the reconstructed local motion is physically grounded on the $SE(3)$ manifold before being mapped, through the above $SE(3)$ calibration, to the corresponding real pose trajectory used for supervision or evaluation. This incremental formulation ensures geometrically consistent motion accumulation and guarantees that the estimated pose remains on the $SE(3)$ manifold throughout the tracking process.


\subsection{Equivariant Tactile Coordinate Consistency}
\label{subsec:consistency}

The output of the estimator is defined in the local tactile frame. This is advantageous because the estimate follows local contact geometry rather than arbitrary appearance variation in the tactile image. In particular, the notion of equivariance in this paper is geometric rather than photometric: the same object motion can produce different tactile images when contact occurs at different sensor locations, but the inferred motion should still remain geometrically consistent under the corresponding change of frame instead of varying arbitrarily with image appearance. Equivariantly, this means that the $SE(3)$ motion estimated from touch is tied to the end-effector frame rather than to the raw tactile appearance. As a result, even if the robot arm changes its global pose or the object makes contact at different sensor locations, the same underlying local object motion should induce the same motion estimate when expressed in the end-effector frame, instead of producing inconsistent outputs due only to shifts in contact location or image appearance. In this sense, the estimator is designed to make the tactile-to-$SE(3)$ mapping invariant in the end-effector coordinates while allowing the raw tactile images themselves to vary across contacts.

This property is practically important because it turns the tactile stream into a control-facing geometric signal. Downstream modules no longer need to interpret raw tactile images directly; they need only consume the estimated translation and rotation in a known sensor-centric frame.




\subsection{Interpretation as a Tactile Geometric Residual}
\label{subsec:interpretation}

\begin{figure}
    \centering
    \includegraphics[width=1\linewidth]{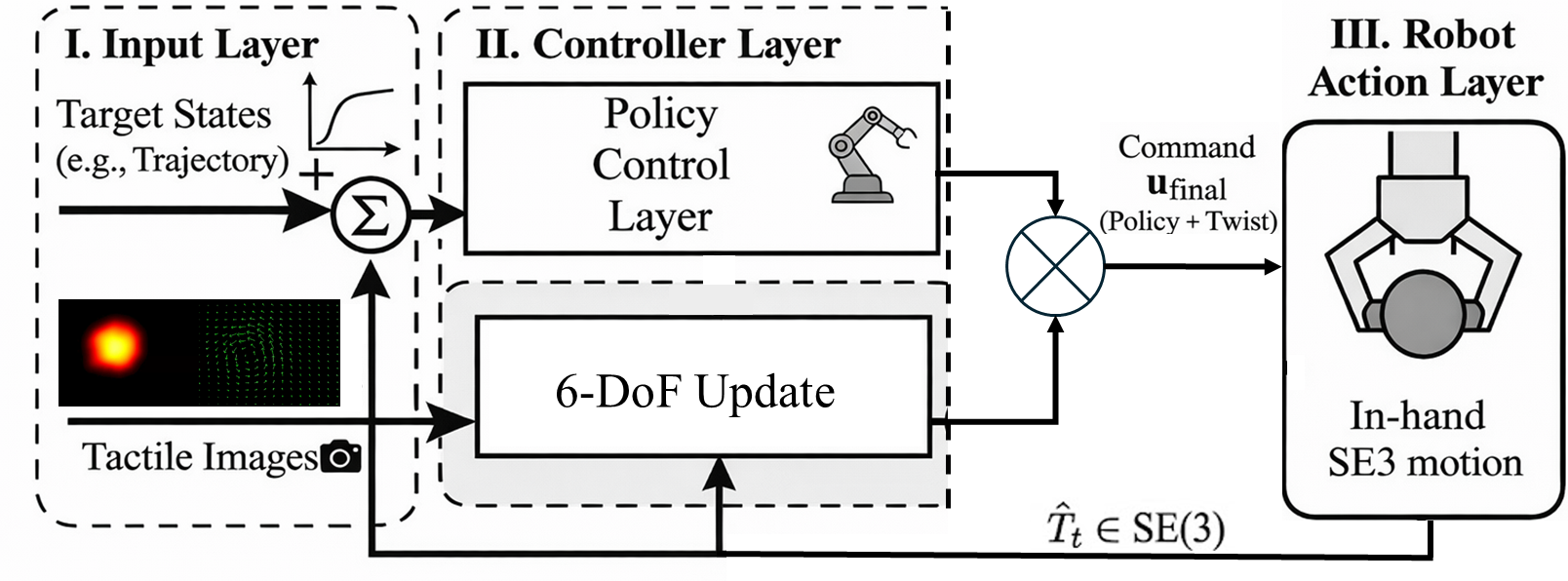}
    \caption{Refined tactile-geometric adjustment in a residual control framework. The tactile-derived $SE(3)$ motion is converted into a 6-DoF update and fused with the policy command to generate the final robot action.}
    \label{fig:execute}
\end{figure}

The estimated $SE(3)$ motion is interpreted as a tactile geometric residual that complements an existing base policy, as illustrated in Fig.~\ref{fig:execute}. Rather than acting as an intermediate latent representation, the predicted twist is converted into a physically interpretable 6-DoF motion update and fused with the policy command to refine the final robot action. This residual formulation enables geometry-aware adjustment without modifying the base-policy structure.

Specifically, the controller layer consists of two parallel components. The policy control layer generates a nominal joint command based on the target state and sensor feedback, while the tactile post-processing layer estimates a 6-DoF geometric update from tactile observations. The two outputs are then combined through a residual fusion module to produce the final command

\begin{equation}
\mathbf{u}_{\text{final}} 
= \mathbf{u}_{\text{policy}} 
+ \Delta \mathbf{u}_{\text{tactile}},
\end{equation}

where $\Delta \mathbf{u}_{\text{tactile}}$ is derived from the estimated $SE(3)$ twist. This formulation preserves the original behavior of the learned policy while enabling real-time tactile-driven geometric correction.

In practice, two forms of geometric information are utilized in the robot action layer. The first is the full pose update $\hat{T}_t \in SE(3)$, which supports online in-hand motion tracking and closed-loop compensation. The second is the rotational component $\hat{R}_t$, or equivalently the incremental rotation extracted from the twist, which can be used as a lightweight adjustment signal for adaptive manipulation.

This residual design allows the tactile geometric module to be seamlessly integrated into existing control pipelines, providing interpretable and physically consistent motion refinement without task-specific retraining.

\section{Experiments and Discussion}
\label{sec:experiments_discussion}

All experiments are conducted on the same robot arm (UR5) and gripper platform (DM-Tac G) so as to isolate the effect of the sensing and estimation method. Two DM-Tac VBTSs from Daimon Robotics are mounted on the opposing gripper fingertips. Each sensor captures tactile signals at 120 Hz with a spatial resolution of $320 \times 240$. During contact, the sensors produce deformation observations that can be decomposed into normal and shear force distributions, which are then used to build the decoupled 3D force field. The dual-sensor configuration is important in practice because it provides complementary contact observations and helps disambiguate cases in which different rigid-body motions induce similar local surface deformation on a single sensor. The evaluation is designed as a controlled validation of the proposed force-field-to-motion mechanism and its downstream utility, rather than as an exhaustive benchmark against all tactile tracking pipelines. This scope is chosen because many correspondence-based baselines rely on visual texture, marker tracks, dense normal-map registration, or stable point-cloud geometry, which are precisely the cues that become unreliable in the low-texture in-gripper setting targeted here.

The implementation details include synchronization with robot state, calibration between the tactile sensor frame and the robot end-effector frame, the optimization frequency, filtering parameters, temporal window size, and the computational hardware used for online inference. Whenever possible, reference rotation measurements are obtained from the robot kinematics, an external motion-capture system, or another calibrated sensing source so that the estimated tactile rotation can be compared against ground truth.

\subsection{Rotation Estimation Accuracy under Controlled Ball Motion}
\label{subsec:exp1}

The experiment evaluates the rotation estimation accuracy when a grasped spherical object undergoes controlled rotational motion. A rigid ball is stably grasped by the gripper, and commanded rotations are applied around the three principal axes of the tactile frame, denoted as $r_x$, $r_y$, and $r_z$. Three target rotation angles, $30^\circ$, $60^\circ$, and $90^\circ$, are selected to evaluate estimation performance under increasing motion magnitude. This setup allows assessment of both axis-wise consistency and error growth under larger rotations.

\begin{figure*}[ht]
    \centering
    \includegraphics[width=1\textwidth]{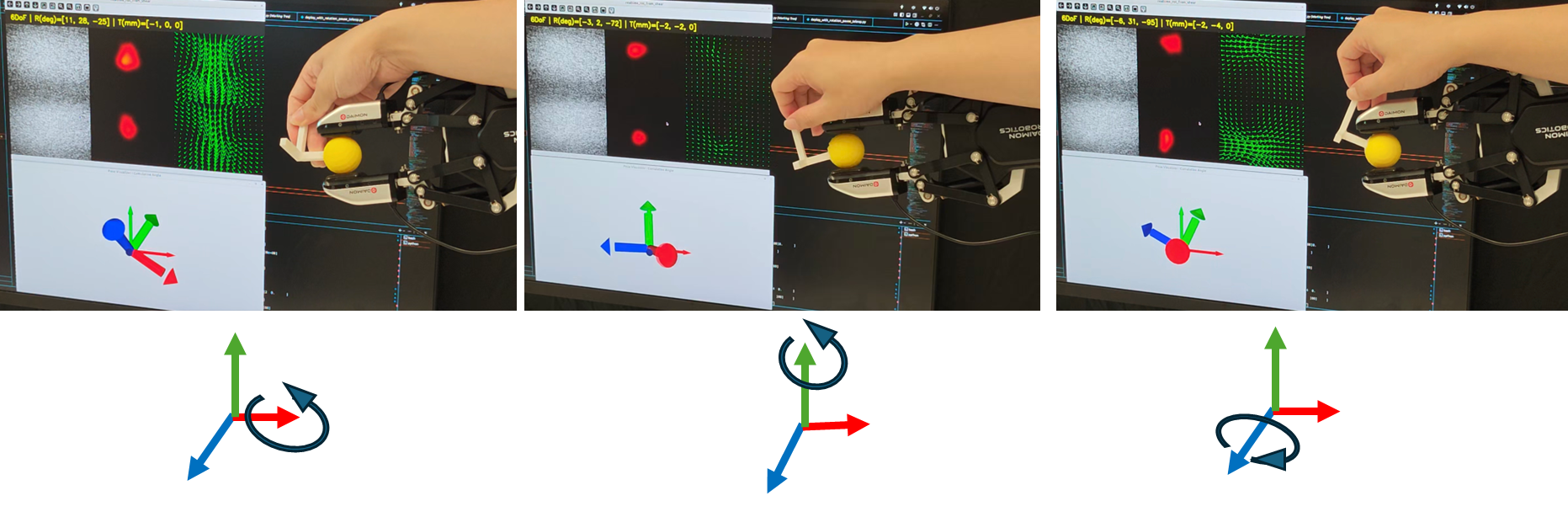}
    \caption{Representative video screenshots of the grasped object rotating about the three principal axes, $r_x$, $r_y$, and $r_z$. The visualization interface shows the tracked 6D pose data together with the tactile image data from the visuotactile sensors.}
    \label{fig:diff_angle}
\end{figure*}

For each trial, the tactile estimator produces an instantaneous angular velocity estimate $\hat{\boldsymbol{\omega}}_t$, which is integrated to obtain the cumulative rotation as described in Section~\ref{subsec:estimate}. The estimated rotation angle is then compared with the commanded ground-truth angle. The evaluation metric is the mean angular error. To reflect realistic sensing behavior, the expected error increases moderately with larger rotation angles due to accumulated integration drift and increased contact nonlinearity. Fig.~\ref{fig:diff_angle} shows representative video screenshots of the object rotating about the three different axes. The visualization interface shows the tracked 6D pose data together with the tactile image data from the visuotactile sensors.

Table~\ref{tab:rotation_accuracy} summarizes the quantitative results. For the smallest rotation of $30^\circ$, the estimation error remains around $3^\circ$, while larger rotations of $60^\circ$ and $90^\circ$ exhibit gradually increasing errors. This behavior is consistent with the short-horizon rigid-body approximation, where larger cumulative rotations introduce more pronounced contact deformation and integration drift.

\begin{table}[ht]
    \centering
    \caption{Rotation estimation accuracy for a grasped ball under controlled rotations across three target angles.}
    \label{tab:rotation_accuracy}
    \begin{tabular}{lccc}
        \toprule
        Axis & Angle ($^\circ$) & Mean Err. ($^\circ$) \\
        \midrule
        $r_x$ & 30 & 3.0 \\
        $r_x$ & 60 & 7.1 \\
        $r_x$ & 90 & 11.9 \\
        \midrule
        $r_y$ & 30 & 3.2 \\
        $r_y$ & 60 & 7.5 \\
        $r_y$ & 90 & 12.2 \\
        \midrule
        $r_z$ & 30 & 2.9 \\
        $r_z$ & 60 & 6.8 \\
        $r_z$ & 90 & 11.7 \\
        \bottomrule
    \end{tabular}
\end{table}

This experiment further validates the robustness of the proposed $SE(3)$-based formulation. Consistent accuracy across $r_x$, $r_y$, and $r_z$ indicates that the method captures the underlying rigid-body rotation rather than relying on axis-specific tactile appearance. This property is particularly important for smooth and symmetric objects without distinct surface textures, where traditional feature-based approaches often fail. The results demonstrate that the force-based $SE(3)$ pose reconstruction achieves reliable accuracy even under such textureless contact conditions.

Moreover, for real robotic manipulation, absolute pose accuracy is not always the most critical factor. Instead, the ability to track incremental rotational changes in real time is often more important for adaptive and dexterous manipulation. The estimated angular variation provides a meaningful feedback signal that can be directly used for closed-loop control and grasp adjustment. By leveraging this real-time rotational feedback, the robot can better compensate for object motion within the gripper, thereby improving manipulation stability and enhancing overall dexterity.

\subsection{Comparison of Single and Dual Sensor Configurations}
\label{subsec:expDual}

\begin{figure*}[ht]
    \centering
    \includegraphics[width=1\linewidth]{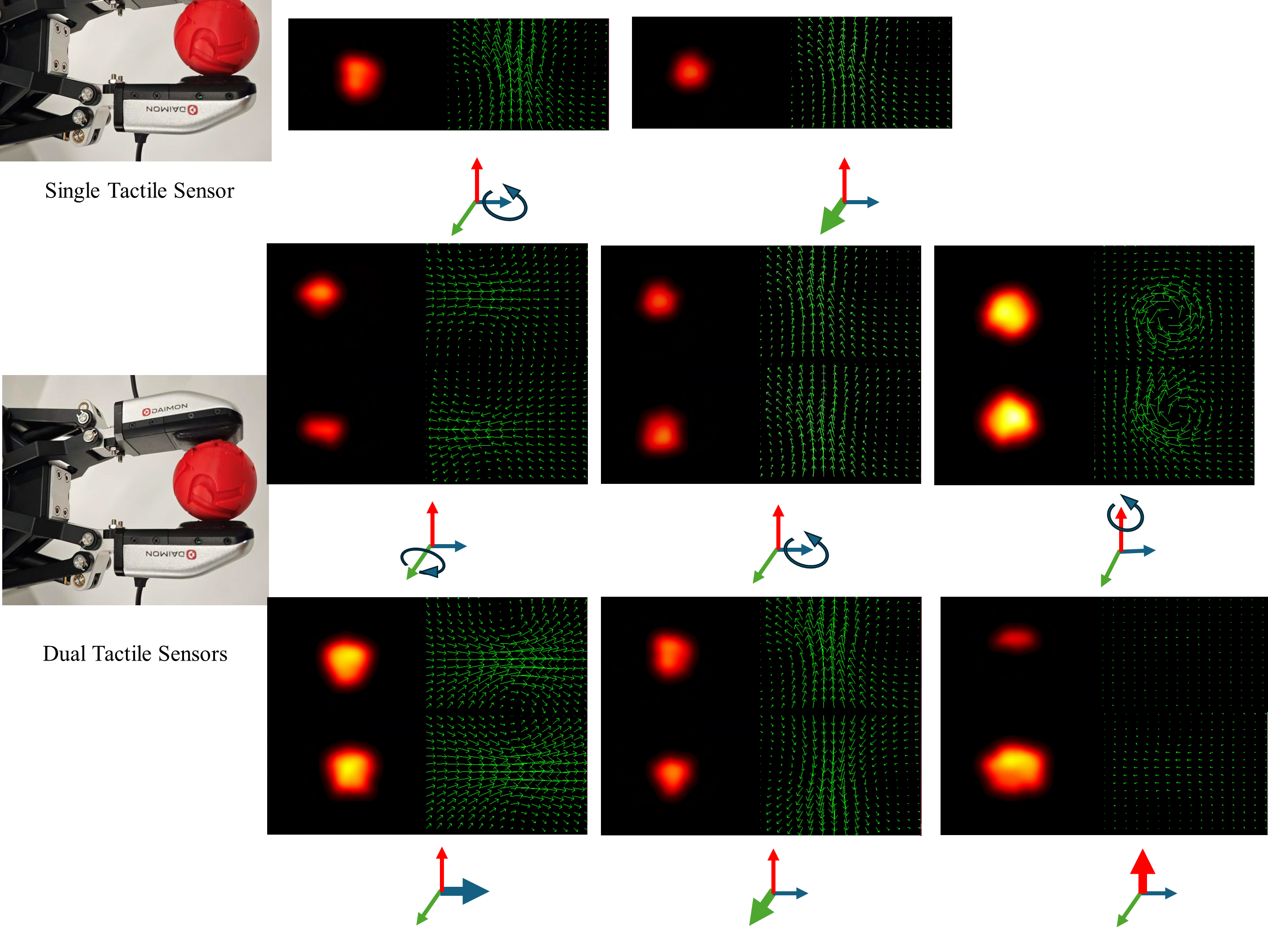}
    \caption{Comparison between single-sensor and dual-sensor configurations. Each subfigure shows the tactile image of the decomposed three-dimensional force field measured by the sensor. The coordinate axes beneath each subfigure indicate the corresponding motion. For the single-sensor configuration, two representative motions are shown, namely rotation about the $x$-axis and translation along the $y$-axis. For the dual-sensor configuration, the visualization covers the full six degrees of freedom.}
    \label{fig:Dualsensor}
\end{figure*}

In this section, we analyze the performance difference between single-sensor and dual-sensor configurations for $SE(3)$ motion estimation. When only a single tactile sensor is used, translational and rotational motions may produce similar 3D force field patterns, leading to ambiguity in motion interpretation. As illustrated in Fig.~\ref{fig:Dualsensor}, both displacement and rotation can generate comparable local force distributions, making it difficult to distinguish between the two motion types. Consequently, the estimated $SE(3)$ pose becomes unreliable, particularly in scenarios involving small rotations or lateral translations.

To address this limitation, we introduce a dual-sensor configuration, where two tactile sensors are placed symmetrically at opposite locations on the gripper. This mirrored placement provides complementary observations of the contact interaction. As shown in Fig.~\ref{fig:Dualsensor}, translational motion produces force fields with consistent directions across the two sensors, whereas rotational motion leads to opposing force patterns. This complementary behavior enables reliable discrimination between displacement and rotation.

Due to the mirror-symmetric placement of the sensors, the coordinate frames of the two tactile sensors exhibit structured sign relationships. Specifically, the $x$-axis directions remain consistent, while the $y$ and $z$ axes are mirrored. As a result, for translational motion along the $y$ direction, the observed force fields appear reversed in the tactile images but correspond to motion in the same physical direction. In contrast, for rotational motion around the $y$ axis, the force fields appear aligned in the tactile images but represent opposite rotational contributions. A similar phenomenon occurs for rotations around the $z$ axis, where the observed force directions on the two sensors are opposite, although the object rotates in the same physical direction. In particular, translation along the $z$ direction increases the normal-force magnitude and effective contact area on one sensor while decreasing them on the other. In typical contact-rich manipulation tasks, however, variation along the $z$ direction is usually small and is therefore neglected in the present analysis.

To fuse the motion estimates from the dual sensors, we combine the estimated twists from each sensor:
\begin{equation}
\boldsymbol{\xi}_i =
\begin{bmatrix}
\boldsymbol{\omega}_i \\
\mathbf{t}_i
\end{bmatrix}
\in \mathbb{R}^6, \quad i \in \{1,2\},
\end{equation}

where $\boldsymbol{\omega}_i$ and $\mathbf{t}_i$ denote the rotational and translational components estimated from the $i$-th sensor, respectively. The fused twist is computed as

\begin{equation}
\boldsymbol{\xi}_{\text{dual}} =
\frac{1}{2}
\left(
\boldsymbol{\xi}_1 +
\mathbf{S} \boldsymbol{\xi}_2
\right),
\end{equation}

where $\mathbf{S}$ is a diagonal sign matrix determined by the relative placement of the sensors:

\begin{equation}
\mathbf{S} =
\text{diag}
\left(
s_{\omega_x},
s_{\omega_y},
s_{\omega_z},
s_{t_x},
s_{t_y},
s_{t_z}
\right).
\end{equation}

For mirror-symmetric placement, the sign vector becomes
$\mathbf{s} = 
\left[
1, -1, -1, 
1, -1, -1
\right]$, 
which accounts for the mirrored coordinate transformation between the two sensors.

To quantitatively evaluate the effectiveness of the dual-sensor configuration, we compare the performance of single-sensor and dual-sensor setups under controlled motion conditions. The experiments include three rotational motions ($R_x$, $R_y$, $R_z$), each with a target rotation of $30^\circ$, and two translational motions ($T_x$, $T_y$), each with a target displacement of 10 mm. We measure the variation of the estimated rotation angles during pure translational motion to assess motion decoupling capability.

\begin{table}[ht]
\centering
\caption{Comparison between single-sensor and dual-sensor configurations. Rotation variation under pure translational motion.}
\label{tab:dual_sensor}
\setlength{\tabcolsep}{4pt}
\begin{tabular}{ccc@{\hspace{4pt}}c}
\toprule
Motion Type & Configuration & Rotation Drift (deg) & Translation Error (mm) \\
\midrule
$T_x$ & Single Sensor & 9.80 & 1.52 \\
$T_x$ & Dual Sensor & 1.10 & 0.64 \\
\midrule
$T_y$ & Single Sensor & 10.60 & 1.72 \\
$T_y$ & Dual Sensor & 1.50 & 0.68 \\
\midrule
$R_x$ & Single Sensor & 7.90 & 1.58 \\
$R_x$ & Dual Sensor & 2.90 & 0.55 \\
\midrule
$R_y$ & Single Sensor & 8.40 & 1.67 \\
$R_y$ & Dual Sensor & 3.20 & 0.57 \\
\midrule
$R_z$ & Single Sensor & 4.90 & 0.74 \\
$R_z$ & Dual Sensor & 3.50 & 0.59 \\
\bottomrule
\end{tabular}
\end{table}

The results demonstrate that the single-sensor configuration frequently produces undesired rotation estimates during pure translational motion, indicating strong coupling between translation and rotation. In contrast, the dual-sensor configuration reduces rotational drift to approximately $3^\circ$, which is consistent with the small-angle ($30^\circ$) rotation-estimation error reported in Table~\ref{tab:rotation_accuracy}. Although non-negligible drift remains, the dual-sensor setup provides substantially better decoupling than the single-sensor baseline and leads to more accurate and stable $SE(3)$ pose estimation.

Overall, the dual-sensor configuration provides improved motion observability and reduced ambiguity in tactile-based motion estimation, enabling more robust and accurate $SE(3)$ tracking.

\subsection{Rotation Estimation Across Objects with Different Geometries}
\label{subsec:DifferentGeo}

\begin{figure}[ht]
    \centering
    \includegraphics[width=0.5\textwidth]{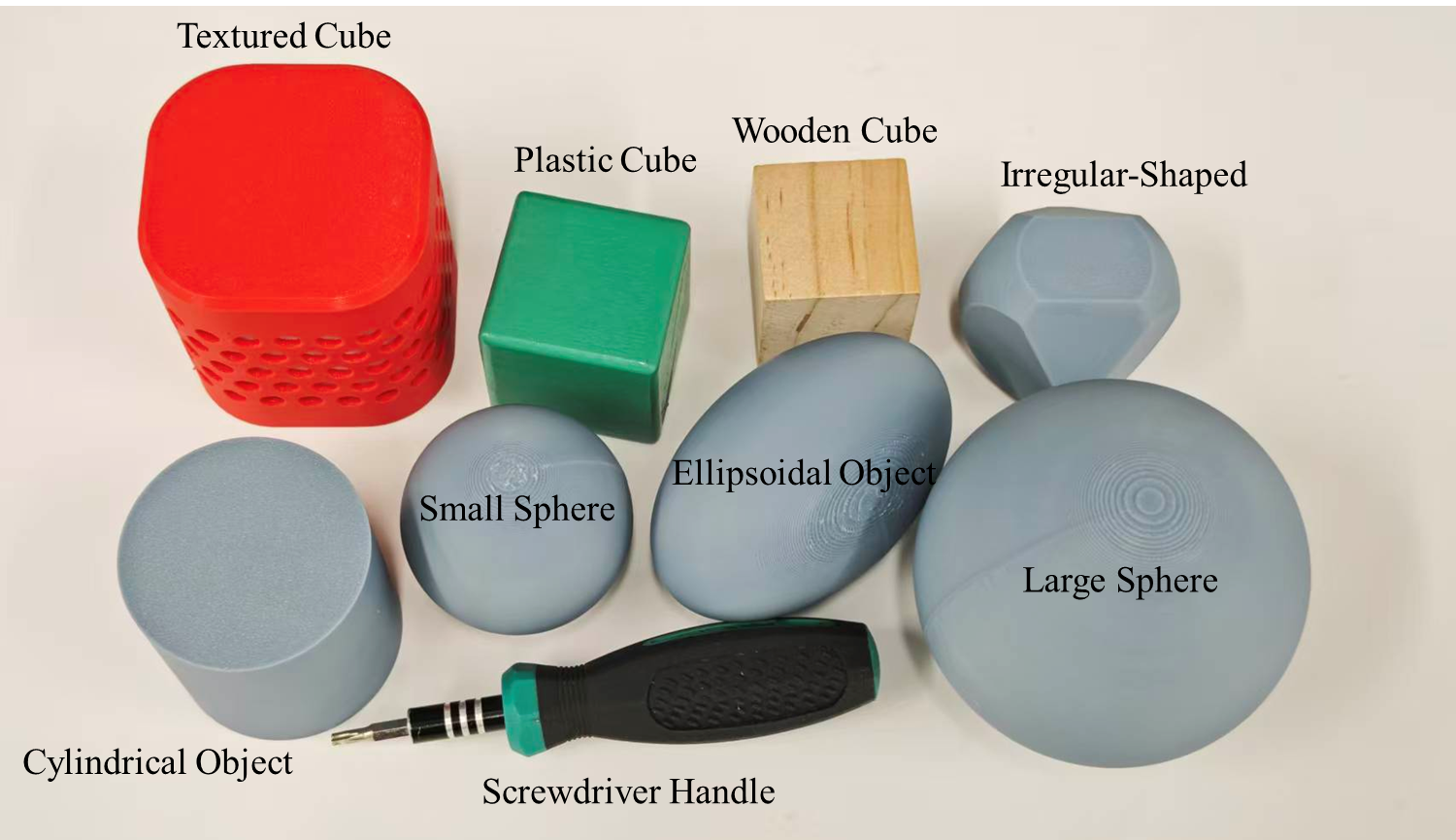}
    \caption{Representative objects used in the multi-object evaluation. Most objects are PLA 3D-printed; the screwdriver handle, wooden cube, and plastic cube provide additional non-PLA cases. Except for the textured cube and the screwdriver handle, the object surfaces are smooth and lack distinctive texture.}
    \label{fig:diff_objects}
\end{figure}

To evaluate generalization beyond a single object category, we further test the proposed tactile rotation estimator on a diverse set of objects that vary primarily in geometry and surface texture, with limited variation in material. As illustrated in Fig.~\ref{fig:diff_objects} and summarized quantitatively in Table~\ref{tab:object_rotation}, most test objects are PLA 3D-printed, including the large sphere, ellipsoidal object, irregular-shaped object, cylindrical object, and textured cube. The main exceptions are the screwdriver handle, the wooden cube, and the plastic cube. In terms of surface appearance, all objects are smooth and textureless except the textured cube and the screwdriver handle. This design allows us to examine whether the method remains effective across different shapes and under both texture-poor and texture-rich tactile conditions, while also providing a limited check across object materials.

\begin{figure*}[ht]
    \centering
    \includegraphics[width=\textwidth]{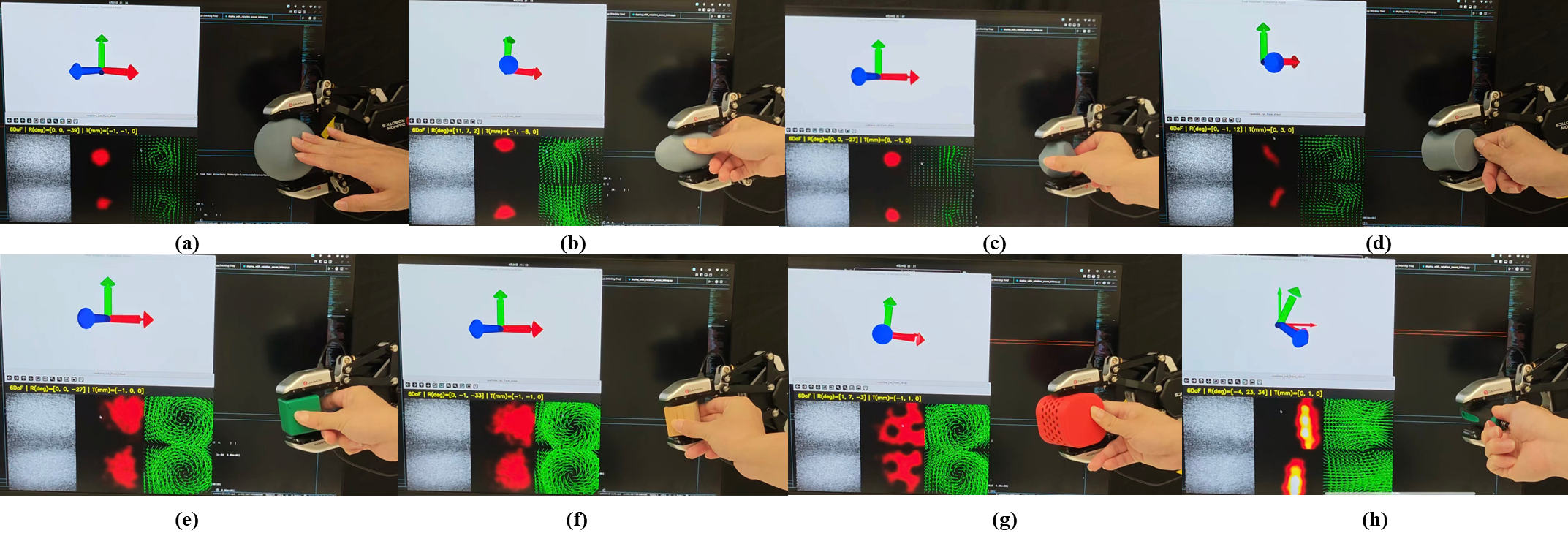}
    \caption{Representative screenshots of the rotation process for the eight objects in the multi-object evaluation. The visualization interface shows the tracked 6D pose data together with the tactile image data from the visuotactile sensors. Subfigures correspond to (a) Large Sphere, (b) Ellipsoidal Object, (c) Irregular-Shaped Object, (d) Cylindrical Object, (e) Plastic Cube, (f) Wooden Cube, (g) Textured Cube, and (h) Screwdriver Handle.}
    \label{fig:diff_rotation}
\end{figure*}

For each object, a controlled rotation of $30^\circ$ is applied under the admissible rotation mode indicated in Table~\ref{tab:object_rotation}. Objects marked as ``Two-axis'' are evaluated under rotations around two observable axes, while objects marked as ``Single-axis'' are evaluated only along the principal axis that yields stable contact. This distinction is imposed by geometric and contact constraints: for some object shapes, stable and informative rotation cannot be induced or observed about all three axes under the grasp configuration used in our experiments. The estimated rotation is then compared with the commanded motion, and the mean angular error over all trials is reported. Fig.~\ref{fig:diff_rotation} visualizes representative screenshots from the rotation process for all eight objects.

The results show consistently low errors across all object categories, ranging from $2.6^\circ$ to $4.0^\circ$. In particular, the method performs well on smooth and weak-texture objects such as the large sphere ($2.9^\circ$), ellipsoidal object ($3.6^\circ$), cylindrical object ($2.8^\circ$), and wooden cube ($3.2^\circ$), indicating that accurate rotation estimation does not depend on distinctive tactile texture. The estimator also remains effective on the two objects with more noticeable surface structure, namely the textured cube ($2.6^\circ$) and the screwdriver handle ($3.5^\circ$). The irregular-shaped object ($4.0^\circ$) and screwdriver handle ($3.5^\circ$) yield slightly larger but still moderate errors, which is reasonable given their more complex contact evolution.

\begin{table}[ht]
    \centering
    \caption{Overall rotation estimation accuracy across objects with different geometries under $30^\circ$ rotation.}
    \label{tab:object_rotation}
    \begin{tabular}{lc}
        \toprule
        Object Type & Mean Angular Error ($^\circ$) \\
        \midrule
        Large Sphere & 2.9 \\
        Ellipsoidal Object & 3.6 \\
        Irregular-Shaped Object & 4.0 \\
        Cylindrical Object (Two-Axis) & 2.8 \\
        Plastic Cube (Single-Axis) & 3.8 \\
        Wooden Cube (Single-Axis) & 3.2 \\
        Textured Cube (Single-Axis) & 2.6 \\
        Screwdriver Handle (Two-Axis) & 3.5 \\
        \bottomrule
    \end{tabular}
\end{table}

These results indicate that the force-based $SE(3)$ reconstruction is robust to variations in both object geometry and surface texture, with preliminary evidence suggesting stability across different materials. Its performance on predominantly smooth, low-texture objects confirms that the method does not rely on salient image correspondences, and its strong results on the textured cube and screwdriver handle show that the same estimator remains applicable when contact appearance becomes more structured. From a manipulation standpoint, this object-agnostic behavior is important because practical in-gripper tasks involve a wide spectrum of geometries and surface conditions. The results therefore support the proposed method as a robust rotational feedback module for diverse real-world objects rather than for a narrowly defined object class.

\subsection{Rotation Output as a Post-Processor under an Existing Policy}
\label{subsec:exp2}

This experiment evaluates whether the estimated rotation can be incorporated into an existing robot policy as a lightweight post-processing signal without modifying the original policy. In this experiment, the base policy is Action Chunking with Transformers (ACT). Expert demonstrations are first collected manually to train the policy, yielding a stable ACT model that serves as the base policy. Let $\mathbf{a}^{\mathrm{base}}_t$ denote the action generated by the base policy at time step $t$. The action is augmented with a tactile correction term derived from the estimated incremental rotation:
\begin{equation}
\mathbf{a}^{\mathrm{corr}}_t 
= 
\mathbf{a}^{\mathrm{base}}_t 
+ 
\Gamma \, \hat{\boldsymbol{\theta}}_t ,
\end{equation}
where $\hat{\boldsymbol{\theta}}_t$ represents the estimated incremental object rotation and $\Gamma$ denotes a mapping matrix that transforms the rotation estimate into a corrective signal in the robot action space. Depending on the manipulation task, this correction can adjust the end-effector orientation, refine grasp alignment, or trigger a local recovery behavior prior to the next control cycle.

\begin{figure}[ht]
    \centering
    \includegraphics[width=0.5\textwidth]{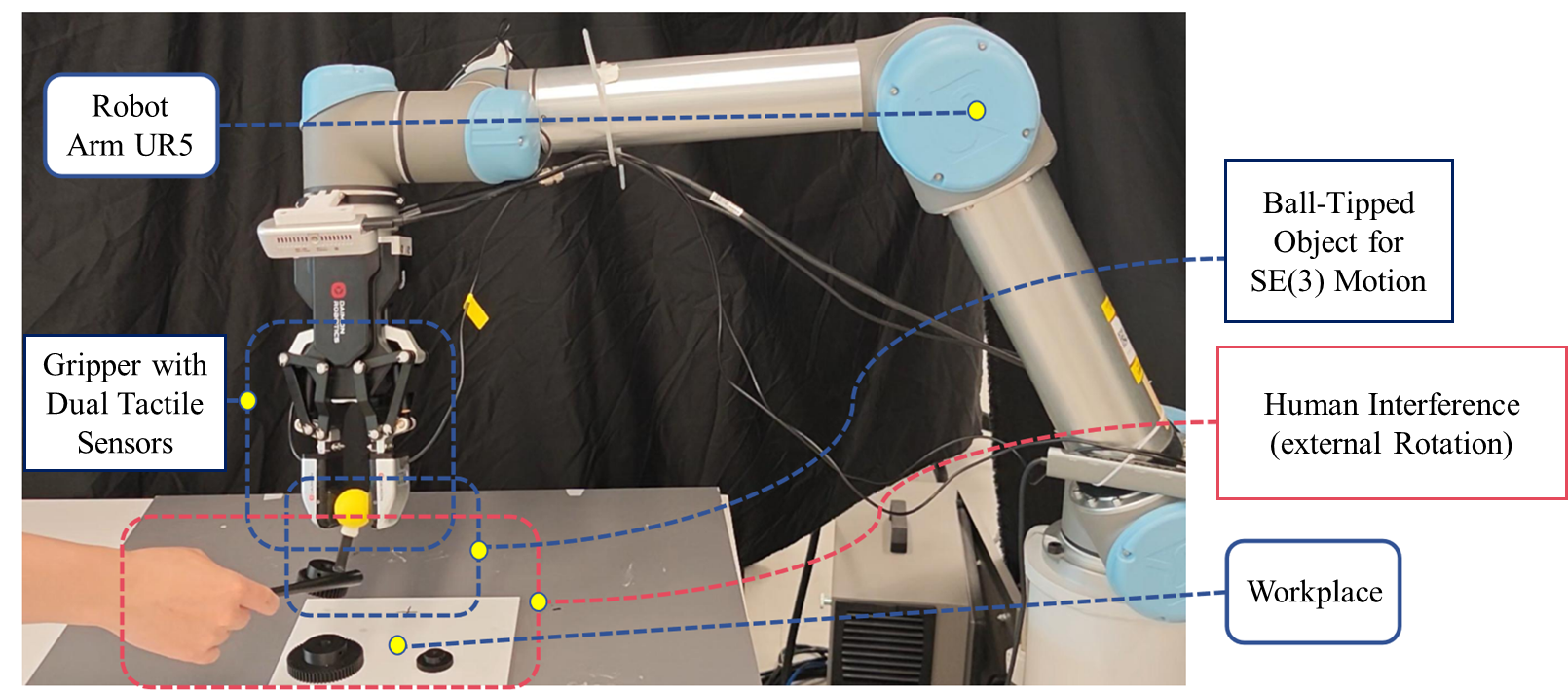}
    \caption{Experimental setup for the policy-level disturbance-recovery study. A UR5 robot arm is equipped with a gripper carrying two visuotactile sensors. During \emph{Human Interference}, an operator applies external rotation to the grasped object. The manipulated tool ends are spherical to facilitate smooth local $SE(3)$ motion during in-gripper interaction.}
    \label{fig:setup}
\end{figure}

To evaluate robustness under external disturbance, we compare three settings: (1) \emph{no human interference} using the base policy alone as a clean reference, (2) \emph{human interference} with the base policy only, and (3) \emph{human interference} with the proposed tactile rotation post-processor enabled. In this setup, human interference means that a human intentionally applies an external rotation to the grasped object during execution, creating in-gripper pose mismatch that the controller must handle online. As shown in Fig.~\ref{fig:setup}, the platform uses a UR5 robot arm equipped with a gripper instrumented with two visuotactile sensors. The manipulated tool ends are spherical, which facilitates smooth local $SE(3)$ motion under external rotation and provides a consistent contact condition for evaluating the proposed compensation module.

\begin{figure*}[ht]
    \centering
    \includegraphics[width=0.85\textwidth]{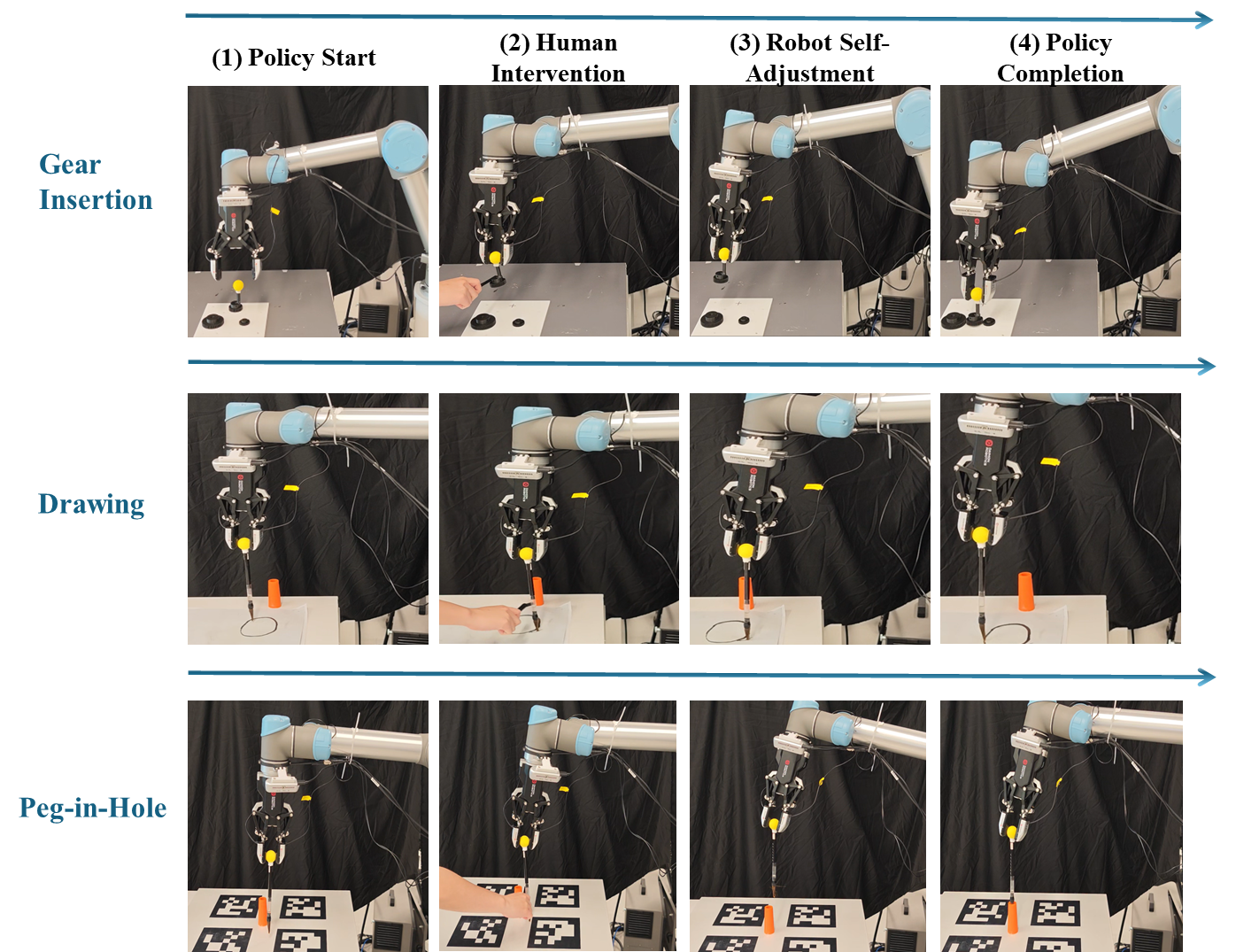}
    \caption{Experimental process for the three policy-level tasks: Drawing, Gear Insertion, and Peg-in-Hole. Each task is illustrated as a sequence of four stages: policy start, human interference, robot self-adjustment, and policy finish.}
    \label{fig:exp_process}
\end{figure*}

The evaluation includes three representative tasks: Drawing (drawing a circle with a brush), Gear Insertion (placing a small gear between two gears), and Peg-in-Hole (inserting the brush into a holder). Each condition is repeated for 20 trials per task. Fig.~\ref{fig:exp_process} summarizes the experimental process for all three tasks. For each task, the sequence proceeds from policy start to human interference, then to robot self-adjustment, and finally to policy finish. Because the disturbance is introduced by physical human interaction, this experiment is treated as a policy-level stress test that reports descriptive success rates rather than a statistically powered benchmark of manipulation policies.
Task-specific success criteria are defined as follows: for Drawing, the brush completes one closed circle with continuous contact and without dropping the object; for Gear Insertion, the small gear is seated in the target slot between the two gears within the allowed time; for Peg-in-Hole, the brush is inserted into the holder.

\begin{table}[ht]
    \centering
    \caption{Task success rate (\%, 20 trials per task) under no-interference and human-interference conditions.}
    \label{tab:post_processor}
    \small
    \setlength{\tabcolsep}{3pt}
    \begin{tabular}{p{3.0cm}ccc}
        \toprule
        Setting  & Gear Insertion & Drawing & Peg-in-Hole \\
        \midrule
        No human interference (Base Policy) & 50\% & 65\% & 75\% \\
        Human interference (Base Policy) & 20\% & 40\% & 25\% \\
        Human interference (Base + Ours) & 35\% & 50\% & 45\% \\
        \bottomrule
    \end{tabular}
\end{table}

Table~\ref{tab:post_processor} summarizes the results. Under human interference, the base policy suffers a substantial performance drop across all three tasks, indicating limited robustness to in-gripper rotational disturbance. After enabling the proposed post-processor, success rates improve in every task, suggesting that the estimated tactile rotation provides actionable feedback for online pose adaptation. Performance with interference also remains below the no-interference reference, which is expected because human perturbation introduces additional uncertainty that cannot be fully eliminated by local correction alone.

\subsection{Integrated Discussion}
\label{subsec:integrated_discussion}

The experiments evaluate the proposed pipeline from three complementary perspectives. The dual-versus single-sensor comparison shows that complementary contacts help reduce translation--rotation ambiguity when motion is inferred from local force fields. Controlled grasped-ball trials then assess intrinsic rotation accuracy across axes and rotation magnitudes, while the multi-object study examines whether the same force-based rotational cue remains informative when contact geometry and surface condition vary. Finally, the policy-level experiment tests whether the estimated rotation remains useful when deployed only as a lightweight post-processing signal. Considering these settings together is important because intrinsic accuracy, cross-condition robustness, and downstream utility need not coincide: a post-processor is only as effective as the reliability of the underlying tactile estimate, and a geometric estimator must remain stable across changing contact conditions.

From a methodological standpoint, the key advantage of the approach is that it treats the tactile stream primarily as a measurement of contact geometry rather than of repeatable image texture. Following the image-to-force structure commonly used in dense-pattern VBTS processing, the tactile observations are organized into a decoupled force field from which rigid motion is inferred. In particular, separating planar translation at the contact centroid from rotation estimated through the shear-related field avoids fitting competing translational and rotational components in a single ill-conditioned least-squares problem. This decoupling yields a rotation estimate with clearer physical meaning and makes it easier to combine information across mirrored fingertips. The policy-level results further show that this rotational signal can improve an existing base policy without retraining the policy itself.

However, the proposed approach has several limitations. The short-horizon rigid-contact assumption can break down under pronounced non-rigid deformation, partial slip, or rapidly changing contact topology. Centroid-based translation also assumes a coherent contact mask; fragmented or highly asymmetric contacts can bias the inferred planar motion. In addition, shear-dominated rotation estimation depends on accurate calibration and on the quality of the normal--tangential decomposition used to construct the force field. More broadly, as with any contact-local module, the estimator is not intended to replace global vision for scene-level state estimation, but rather to complement it in contact-rich and occlusion-prone situations.

Future work will focus on making the tactile compensation module more deployable in closed-loop manipulation. Important directions include feeding the estimated motion directly into policy inputs, learning lightweight corrective-action models from the force-field representation, and evaluating long-horizon robustness across a broader range of tools, grasps, and contact conditions. We also plan to develop confidence weighting and contact gating for slipping, noisy, edge, or disconnected contacts, enabling more reliable fusion across multiple fingertips and tighter integration with vision or known object geometry.

\section{Conclusion}
\label{sec:conclusion}

This paper presented a method for estimating rigid-body motion on $SE(3)$ from low-texture visuotactile images through a decoupled three-dimensional force field and manifold-consistent twist integration. Following dense-pattern tactile processing, the method extracts normal and tangential responses, derives planar translation from the motion of the contact-mask centroid on the sensor plane, and estimates rotation from the shear-related field without jointly fitting an additional uniform in-plane translation in the same least-squares problem. This centroid--shear decoupling provides a clear physical interpretation of contact-patch motion and shear-field rotation, while also improving numerical conditioning relative to a fully coupled translation--rotation fit.

Experiments on paired DM-Tac sensors showed that dual-finger sensing reduces translation--rotation ambiguity, controlled grasped-ball trials provided reliable rotation tracking across axes and rotation magnitudes, and multi-object evaluation supported generalization across different geometries and surface conditions. In addition, policy-level experiments showed that the estimated rotation can serve as an effective post-processing signal to improve disturbance tolerance without modifying the base policy itself. Overall, these results support force-field-to-$SE(3)$ estimation with explicit translation--rotation decoupling as a practical geometric bridge between low-texture visuotactile sensing and contact-rich in-gripper manipulation.

\section*{acknowledgment}

The author extends sincere appreciation to the Hong Kong Center for Construction Robotics (InnoHK center supported by the Hong Kong ITC, InnoHK-HKCRC) for the funding support.

\bibliographystyle{IEEEtran}
\bibliography{reference}

\end{document}